%% file: Camera_ready.tex
  \documentclass[sigconf]{acmart}

\usepackage{pifont} % 提供 \ding 命令
\usepackage{subcaption}
\usepackage{multirow}
\usepackage{float}
\usepackage{bm}
\usepackage{colortbl}
\usepackage[table]{xcolor}
\usepackage{marvosym}

\usepackage[perpage,symbol*]{footmisc}
\DefineFNsymbols{circled}{{\ding{192}}{\ding{193}}{\ding{194}}
{\ding{195}}{\ding{196}}{\ding{197}}{\ding{198}}{\ding{199}}{\ding{200}}{\ding{201}}}
\setfnsymbol{circled}

\newcommand\myfootnotestyle[1]{\ifcase#1 \or \ding{182}\or \ding{183}\or
\ding{184}\or \ding{185}\or \ding{186}\or \ding{187}%
\or \ding{188}\or \ding{189}\or \ding{190}\or \ding{191}\else *\fi\relax}
\newcommand{\ie}{\textit{i}.\textit{e}.}
\newcommand{\eg}{\textit{e}.\textit{g}.}

\newcommand{\Fref}[1]{Fig.~\ref{#1}}

\newcommand{\toolns}{\textsc{MetAdv}}
\newcommand{\tool}{\toolns\space}

\AtBeginDocument{%
  }

\copyrightyear{2025}
\acmYear{2025}
\setcopyright{rightsretained}
\acmConference[MM '25]{Proceedings of the 33rd ACM International Conference on Multimedia}{October 27--31, 2025}{Dublin, Ireland}
\acmBooktitle{Proceedings of the 33rd ACM International Conference on Multimedia (MM '25), October 27--31, 2025, Dublin, Ireland}
\acmDOI{10.1145/3746027.3754477}
\acmISBN{979-8-4007-2035-2/2025/10}

\begin{document}

\title{\toolns: A Unified and Interactive Adversarial Testing \\Platform for Autonomous Driving}
%\title{DualInject: Dual-Modal Prompt Injection Attack against Large Multimodal Agents}

 %\email{webmaster@marysville-ohio.com}
 %\affiliation{%
   %\institution{Institute for Clarity in Documentation}

% \affiliation{%

% \affiliation{%

% \affiliation{%

% \affiliation{%
\author{Aishan Liu}
\affiliation{%
  \institution{Beihang University, Beijing, China}
  \city{}
  \country{}
}
\email{liuaishan@buaa.edu.cn}
\orcid{0000-0002-4224-1318}

\author{Jiakai Wang}
\affiliation{%
  \institution{Zhongguancun Laboratory, Beijing, China}
  \city{}
  \country{}
}
\email{wangjk@zgclab.edu.cn}
\orcid{}

\author{Tianyuan Zhang}
\affiliation{
  \institution{Beihang University, Beijing, China}
  \city{}
  \country{}
}
\email{zhangtianyuan@buaa.edu.cn}
\orcid{0000-0001-9874-6828}

\author{Hainan Li}
\affiliation{%
  \institution{Institute of Dataspace, Hefei, China}
  % \city{Hefei}
  \city{}
  \country{}
}
\email{hainan@buaa.edu.cn}
\orcid{0009-0002-6638-8282}

\author{Jiangfan Liu}
\affiliation{%
  \institution{Beihang University, Beijing, China}
  % \city{Hefei}
  \city{}
  \country{}
}
\email{liujiangfan@buaa.edu.cn}
\orcid{}

\author{Siyuan Liang}
\affiliation{%
  \institution{Nanyang Technological University, Singapore, Singapore}
  \city{}
  \country{}
}
\email{pandaliang521@gmail.com}
\orcid{0000-0002-6154-0233}

\author{Yilong Ren}
\affiliation{%
  \institution{Beihang University, Beijing, China}
  % \city{Hefei}
  \city{}
  \country{}
}
\email{yilongren@buaa.edu.cn}
\orcid{}

\author{Xianglong Liu\dag}
\affiliation{%
  \institution{Beihang University, Beijing, China}
  % \city{Beijing}
  \city{}
  \country{}
}
\affiliation{%
  \institution{Zhongguancun Laboratory, Beijing, China}
  % \city{Beijing}
  \city{}
  \country{}
}
\affiliation{%
  \institution{Institute of Dataspace, Hefei, China}
  % \city{Hefei}
  \city{}
  \country{}
}
\email{xlliu@buaa.edu.cn}
\orcid{0000-0001-8425-4195}

\author{Dacheng Tao}
\affiliation{%
  \institution{Nanyang Technological University, Singapore, Singapore}
  \city{}
  \country{}
}
\email{dacheng.tao@ntu.edu.sg}
\orcid{}

\thanks{\dag The Corresponding author.}

% \affiliation{%

% \affiliation{%

\renewcommand{\shortauthors}{Aishan Liu et al.}


\input{secs/0_abstract}

\maketitle

\begin{figure}
    \centering
    \vspace{-0.15in}
    \includegraphics[width=1.00\linewidth]{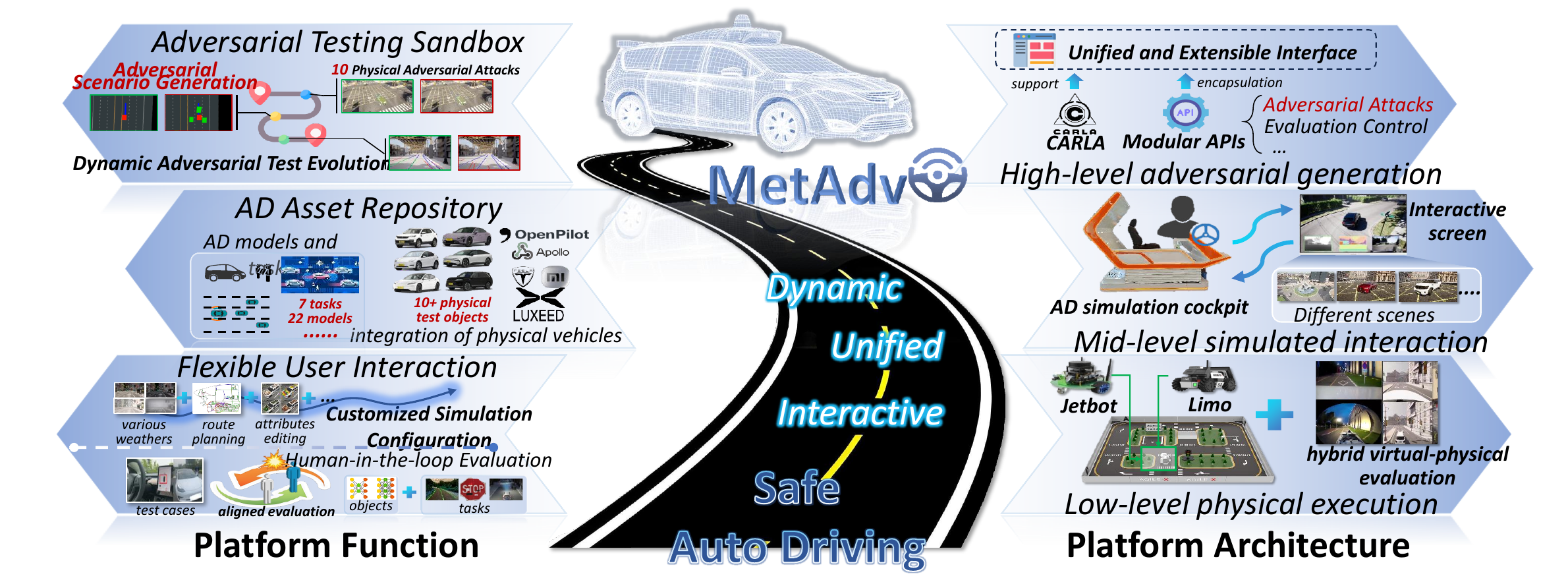}
    \vspace{-0.35in}
    \caption{The overall framework of our \toolns, a novel virtual-physical hybrid adversarial testing platform for AD.}
    \label{fig:framework}
    \vspace{-0.15in}
\end{figure}

\section{Introduction}

Autonomous driving has made significant progress in recent years \cite{zimmer20193d, lang2019pointpillars, erccelik20223d, zimmer2022real, li2024bevformer, hu2023_uniad, chen2024end, jiang2023vad}, yet remain highly susceptible to \emph{adversarial attacks} \cite{szegedy2013intriguing, FGSM}, ranging from digital perturbations in sensor input \cite{dong2023benchmarking, hallyburton2022security} to physical-world camouflage \cite{eykholt2018robust, liu2019perceptual, wang2021dual}, which can lead to critical failures and safety hazards. As these systems move closer to widespread deployment, evaluating and ensuring their adversarial robustness has become an urgent and unresolved challenge. However, current adversarial testing approaches are hampered by the lack of a unified, interactive, and dynamic evaluation platform. Existing studies \cite{xu2022safebench, li2022metadrive} often focus narrowly on isolated and static environments, limiting their capacity to uncover the full spectrum of weaknesses that may emerge in real-world conditions. 

To address this, we present \toolns, a novel, unified, and interactive adversarial testing platform for AD (shown in \Fref{fig:framework}). Central to \tool is a hybrid virtual-physical sandbox built on CARLA, which enables real-time, closed-loop adversarial evaluation from virtual simulation to physical deployment. \tool features a three-layer hierarchical architecture, unifying the high-level adversarial attack generation and connecting mid-level simulation operations to low-level physical execution. Here, we also propose dynamic adversarial testing evolution that can generate targeted adversarial test cases to better reveal threats covering the full perception, prediction, and planning pipeline. \tool is further distinguished by its integration of a rich AD asset repository, encompassing a broad spectrum of tasks, algorithmic paradigms, and 3D vehicle models. It also supports seamless transitions from simulation to deployment, with plug-and-play compatibility for commercial platforms such as Apollo \cite{fan2018baidu, fan2018auto} and Tesla \cite{tesla_ai_website}. Besides, \tool offers extensive user interactivity, allowing real-time editing of environments, adversarial artifact insertion; crucially, it supports human-in-the-loop evaluation: by capturing real-time physiological signals and behavioral feedback (\eg, electroencephalography), it enables the study of human-machine trust under adversarial conditions. \tool supports software-in-the-loop (SIL), hardware-in-the-loop (HIL), and vehicle-in-the-loop (VIL) testing, which serve as a diagnostic tool to expose hidden vulnerabilities for more trustworthy AD vehicles.

%Meanwhile, \tool integrates a rich \ding{183} AD asset repository, containing a diverse collection of AD algorithms (\eg, traditional deep learning methods, end-to-end learning approaches, and emerging large language model-based driving agents), tasks, and 3D models; additionally, \tool achieves seamless transition from virtual simulation to real-world vehicle execution with plug-and-play interfaces for commercial AD platforms such as Apollo \cite{?} and Tesla \cite{?}. Finally, \tool enables \ding{184} flexible user interaction, where users can also flexibly modify simulation scenes, insert adversarial artifacts, and choose among different models for direct comparison and robustness evaluation; importantly, \tool allows human-in-the-loop evaluation, where driver behavior and feedback (\eg, electroencephalography) can be captured in real-time to further calibrate and study for human-machine mutual trust under adversarial conditions. Our platform supports software-in-the-loop (SIL), hardware-in-the-loop (HIL), and vehicle-in-the-loop (VIL) testing, which can serve as a diagnostic tool to expose hidden vulnerabilities for more trustworthy autonomous vehicles. 

\section{\tool System}
\subsection{Adversarial Testing Sandbox}
%To build a dynamic and interactive testing sandbox, 
We first propose a unified closed-loop evaluation framework that enables the real-time adversarial testing and bridges the gap between simulation and reality; we then propose the dynamic adversarial testing evolution that generate adversarial cases targeted to AD.% to reveal the potential threats covering the perception-prediction-planning pipeline.

\textbf{Unified Closed-loop Evaluation.}
The overall framework of our system consists of three hierarchical layers:
\ding{182} \textit{High-level Unified API Layer.} Built upon CARLA 0.9.15 \cite{dosovitskiy2017carla}, this layer provides a unified and extensible interface for adversarial evaluation. We encapsulate core functionalities (\eg, adversarial attacks, evaluation control) into modular APIs to enable standardized and reusable testing workflows. \ding{183} \textit{Mid-level Simulation Operation Layer.} This layer manages interaction with the simulator, including scenario deployment, vehicle control, and sensor emulation. It supports precise control over environmental dynamics and agent behaviors. \ding{184} \textit{Low-level Physical Execution Layer.} This layer enables the connection of simulation control commands to real-world physical vehicles via hardware integration in a faithfully reproduced setting. This framework enables closed-loop evaluation in the interactive environment, supporting SIL, HIL, and VIL level testing.

\textbf{Dynamic Adversarial Test Evolution.} 
Based on the evaluation environment, we then propose to generate adversarial test cases targeted at the test-taking AD. In particular, \tool can generate dynamic \emph{adversarial scenario} to the AD to impact their decision: given a initial benign scene, we use a large language model (LLM) to infer an adversarial agent whose behavior poses a threat to the ego AD; subsequently, we perform scenario evolution by introducing background vehicles with collaborative risky trajectories to increase the likelihood of collision. Based on the adversarial scenario, we can also perform targeted \emph{digital} and \emph{physical adversarial attacks} to the ego AD. Our system features 23 prominent digital adversarial attacks (\eg, FGSM \cite{FGSM}, PGD \cite{PGD}) to rigorously evaluate the robustness of perception modules in ADs by adding imperceptible perturbations to raw sensor data. Besides, \tool has implemented 10 physical adversarial attacks, \eg, DAS \cite{wang2021dual} and FCA \cite{wang2022fca}, that applying adversarial camouflage onto physical objects, using a novel dual-renderer fusion strategy (a differentiable renderer to simulate the adversarial textures and a native CARLA renderer to preserving the unperturbed background ).

\subsection{AD Asset Repository} %@hn
%To enable comprehensive and convenient evaluation, we introduce a rich AD asset repository (\eg, algorithms, tasks, vehicles).

\textbf{AD Models and Tasks.}
\tool supports a diverse range of AD algorithms covering perception, prediction, and planning stages, and includes 7 classical AD tasks (\eg, obstacle recognition, scene understanding). Specifically, \tool consists of 10 deep learning based perception models (\eg, YOLOs \cite{yolo}, SMOKE \cite{smoke}), 5 reinforcement learning based decision AD models (\eg, DDPG \cite{DDPG}, PPO \cite{PPO}). Additionally, we incorporate unified frameworks that cover the full pipeline, including 4 end-to-end models (\eg, UniAD \cite{hu2023planning}, VAD \cite{jiang2023vad}) and 3 LLM-based AD models (\eg, LMDrive \cite{shao2024lmdrive}, Dolphins \cite{ma2023dolphins}). We also introduce 6 high-quality 3D vehicle models represented in .obj files, such as Tesla Model 3.  

\textbf{Integration of Physical Vehicles.} Besides the SIL test, we enable our system with HIL and VIL tests and endow \tool with a progressive testing ability for the integration of physical hardware and vehicles to achieve the practical AD adversarial testing. Benefited from the integrated progressive verification system, \tool has provided the adversarial testing ability on 2 AD platforms (Apollo \cite{fan2018baidu, fan2018auto}, OpenPilot \cite{chen2022level}), 2 laboratory vehicles (LIMO \cite{pimenta2025limo}, Jetbot \cite{nvidia_jetbot}), and 5 commercial vehicles (Tesla Model 3/Y \cite{tesla_model3, tesla_modely}, XIAOMI SU7 \cite{xiaomi_su7}, XPENG G6/P7+ \cite{xpeng_g6, xpeng_p7plus}, and LUXEED \cite{luxeed}). %Moreover, it is necessary to highlight that \tool has the resilient potential and could easily adapt to more physical vehicles.

 %Specifically, \tool provides the 3-layer testing framework, namely, the software-in-the-loop (SIL) test, the hardware-in-the-loop (HIL) test, and the vehicle-in-the-loop (VIL) test, for users to achieve the goal of accurate, effective, and practical AD adversarial robustness testing.

%and 5 integrated perception–decision AD models, encompassing both large-scale models and end-to-end architectures. The perception module primarily includes three functional components: obstacle detection, lane detection, and traffic sign recognition.

%Obstacle detection comprises 2D and monocular 3D object detection models, focusing on dynamic and static objects such as vehicles, pedestrians, and construction barriers. Lane detection provides structural road boundaries and directional cues essential for motion planning and vehicle control. Traffic sign recognition ensures adherence to traffic regulations, forming the basis for lawful autonomous driving.

%Beyond modular perception models, \tool incorporates advanced unified frameworks such as Garage, UniAD, VAD, Interfuser and LMDrive for autonomous driving. These models enable direct mapping from sensory input to control commands via holistic scene understanding. Compared to traditional pipeline-based systems, integrated architectures have emerged as a research focus due to their advantages in joint multi-task optimization and lossless information propagation throughout the processing pipeline.

\subsection{Flexible User Interaction} 

\textbf{Customized Simulation Configuration.} \tool offers a highly customizable environment for evaluating ADs via a set of unified interfaces. Environmental settings such as cloud density, solar position, and fog concentration can be precisely adjusted to emulate various weathers; route planning is supported through the specification of start and end coordinates with a diverse set of map layouts; vehicle attributes, including type and texture, are also configurable to simulate variability in visual appearance; attack methods, AD models, and tasks can be easily configured. Our system also supports the easy extension of new attacks, algorithms, and maps to better simulate the real-world AD conditions.

\textbf{Human-in-the-loop Evaluation.} \tool also involves human-in-the-loop evaluation and considers the more practical scenario where humans and machines work collaboratively in the real world. Specifically, \tool collects real-time human feedback (\ie, electroencephalography, eye movement) and enables drivers participating in the whole driving process for a trustworthy AD evaluation at both HIL and VIL. Specifically, at HIL level, \tool is been deployed on an AD simulation cockpit which allows the users to participate in the test flow and interrupt the operation of the tested AD system, thus providing a human-machine aligned perspective study; at VIL level, \tool offers a real-vehicle monitoring environment to provide a more humane decision-making evaluation dimension. Every takeover by human drivers could be recorded in this environment, allowing users to intuitively judge the adversarial robustness of an AD system by the comparative analysis of takeover frequencies. This design has great potential to be explored, especially for AD testing of human-machine mutual trust.

\section{Acknowledgments}
This work was supported by the National Natural Science Foundation of China (Grant. 62206009), State Key Laboratory of Complex \& Critical Software Environment (CCSE), Aeronautical Science Fund (Grant. 20230017051001), the Fundamental Research Funds for the Central Universities, and the Outstanding Research Project of Shen Yuan Honors College, BUAA (Grant. 230123206).

%\section{Conclusion}

% \newpage
\bibliographystyle{ACM-Reference-Format}
\balance
\bibliography{sample-base}

\end{document}

%% file: secs/0_abstract.tex
\begin{abstract}

Evaluating and ensuring the adversarial robustness of autonomous driving (AD) systems is a critical and unresolved challenge. This paper introduces \toolns, a novel adversarial testing platform that enables realistic, dynamic, and interactive evaluation by tightly integrating virtual simulation with physical vehicle feedback. At its core, \tool establishes a hybrid virtual-physical sandbox, within which we design a three-layer closed-loop testing environment with dynamic adversarial test evolution. This architecture facilitates end-to-end adversarial evaluation, ranging from high-level unified adversarial generation, through mid-level simulation-based interaction, to low-level execution on physical vehicles. Additionally, \tool supports a broad spectrum of AD tasks, algorithmic paradigms (e.g., modular deep learning pipelines, end-to-end learning, vision-language models). It supports flexible 3D vehicle modeling and seamless transitions between simulated and physical environments, with built-in compatibility for commercial platforms such as Apollo and Tesla. A key feature of \tool is its human-in-the-loop capability: besides flexible environmental configuration for more customized evaluation, it enables real-time capture of physiological signals and behavioral feedback from drivers, offering new insights into human-machine trust under adversarial conditions. We believe \tool can offer a scalable and unified framework for adversarial assessment, paving the way for safer AD. Our demo can be found at \url{https://sites.google.com/view/metadv-demo-video}.

%Our system highlights several key capabilities: (1) interactive attack deployment and visualization in real-time; (2) smooth transition from virtual simulation to physical vehicle execution; (3) dynamic performance diagnostics across various attack types and driving scenarios. 

%closed-loop adversarial evaluation platform that integrates virtual simulation with physical vehicle feedback to enable full-stack security evaluation for AD systems. 

\end{abstract}